\pdfoutput=1
\RequirePackage{fix-cm}
\documentclass[smallcondensed,natbib]{svjour3}       
\smartqed  
\usepackage{graphicx}
\usepackage{mathtools}

\usepackage{amsmath}
\usepackage{xcolor}
\usepackage{hyperref}

\newcommand{\changed}[1]{{#1}}

\newcommand{\quotedtext}[1]{{\emph{\textquoteleft {#1}\textquoteright}}}
\begin{document}

\title{How Important is Syntactic Parsing Accuracy? An Empirical Evaluation on Rule-Based Sentiment Analysis\thanks{Carlos G{\'o}mez-Rodr{\'i}guez has received funding from the European Research Council (ERC), under the European Union's Horizon 2020 research and innovation programme (FASTPARSE, grant agreement No 714150), Ministerio de Econom\'{i}a y Competitividad (FFI2014-51978-C2-2-R), and the Oportunius Program (Xunta de Galicia). Iago Alonso-Alonso was funded by an Oportunius Program Grant (Xunta de Galicia). David Vilares has received funding from the Ministerio de Educaci\'{o}n, Cultura y Deporte (FPU13/01180) and Ministerio de Econom\'{i}a y Competitividad (FFI2014-51978-C2-2-R).}
}



\titlerunning{How Important is Syntactic Parsing Accuracy?}        

\author{Carlos G\'{o}mez-Rodr\'{\i}guez \and Iago Alonso-Alonso \and David Vilares 
}


\institute{ Carlos G\'{o}mez-Rodr\'{\i}guez \and Iago Alonso-Alonso \and David Vilares \at
              FASTPARSE Lab, Grupo LyS, Departamento de Computaci\'{o}n, Universidade da Coru\~{n}a \\
	      			Campus de A Coru\~{n}a s/n, 15071, A Coru\~{n}a, Spain \\
              Tel.: +34 881 01 1396\\
              Fax: +34 981 167 160\\
              \email{carlos.gomez@udc.es, iago.alonso@udc.es, david.vilares@udc.es}           
}

\date{Received: date / Accepted: date}

\maketitle

\vspace{-0.7cm}
\noindent\fbox{%
    \parbox{\textwidth}{%
        \small This is the accepted manuscript (final peer-reviewed manuscript) accepted for publication in Artificial Intelligence
        Review, and may not reflect subsequent changes resulting from the publishing process such as editing, formatting,  pagination, and other quality control mechanisms. The final publication in Artificial Intelligence Review is available at link.springer.com via \url{http://dx.doi.org/10.1007/s10462-017-9584-0}.
    }%
}
\vspace{0.2cm}

\begin{abstract}


Syntactic parsing, the process of obtaining the internal structure of sentences in natural languages, is a crucial task for artificial intelligence applications that need to extract meaning from natural language text or speech. Sentiment analysis is one example of application for which parsing has recently proven useful.

In recent years, there have been significant advances in the accuracy of parsing algorithms. In this article, we perform an empirical, task-oriented evaluation to determine how parsing accuracy influences the performance of a state-of-the-art rule-based sentiment analysis system that determines the polarity of sentences from their parse trees. In particular, we evaluate the system using four well-known dependency parsers, including both current models with state-of-the-art accuracy and more innacurate models which, however, require less computational resources.

The experiments show that all of the parsers produce similarly good results in the sentiment analysis task, without their accuracy having any relevant influence on the results. Since parsing is currently a task with a relatively high computational cost that varies strongly between algorithms, this suggests that sentiment analysis researchers and users should prioritize speed over accuracy when choosing a parser; and parsing researchers should investigate models that improve speed further, even at some cost to accuracy.




\end{abstract}

\section{Introduction}
\label{intro}

Having computers successfully understand the meaning of sentences in human languages is a long-standing key goal in artificial intelligence (AI). While full understanding is still far away, recent advances in the field of natural language processing (NLP) have made it possible to implement systems that can successfully extract relevant information from natural language text or speech. Syntactic parsing, the task of finding the internal structure of a sentence, is a key step in that process, as the predicate-argument structure of sentences encodes crucial information to understand their semantics. For example, a text mining system that needs to generate a report on customers' opinions about phones may find statements like ``the iPhone is much better than the HTC 10'' and ``the HTC 10 is much better than the iPhone'', which are identical in terms of the individual words that they contain. It is the syntactic structure -- in this case, the subject and the attribute of the verb to be -- that tells us which of 
the phones is preferred by the customer.

In recent years, parsing has gone from a merely promising basic research field to see widespread use in useful AI applications such as machine translation \citep{Miceli2015translation,Xiao2016translation}, information extraction \citep{Song2015relex,Yu2015relex}, textual entailment recognition \citep{Pado2015te}, learning for game AI agents \citep{Branavan2012} or sentiment analysis \citep{JosPen2009a,VilAloGomNLE2015,VilAloGomJIS2015}. Meanwhile, researchers have produced improvements in parsing algorithms and models that have increased their accuracy, up to a point where some parsers have achieved levels comparable to agreement between experts on English newswire text \citep{Berzac2016bias}, although this does not generalize to languages that present extra challenges for parsing \citep{Farghaly2009arabic} or to noisy text such as tweets \citep{Kong2014tweets}. However, parsers consume significant computational resources, which can be an important concern in large-scale applications \citep{
Clark2009large}, and the most accurate models often come at a higher computational cost \citep{Andor2016face,Gom2016restricted}. Therefore, an interesting question is how much influence parsing accuracy has on the performance of downstream applications, as this can be essential to make an informed choice of a parser to integrate in a given system. 

In this article, we analyze this issue for sentiment analysis (SA), i.e., the use of natural language processing to extract and identify subjective information (opinions about relevant entities) from natural language texts. Sentiment analysis is one of the most relevant practical applications of NLP, it has been recently shown to benefit from parsing \citep{SocPerWuChuManNgPot2013a,VilAloGomNLE2015} and it is especially useful at a large scale (as millions of texts of potential interest for opinion extraction are generated every day in social networks), making the potential accuracy vs. speed tradeoff especially relevant.

For this purpose, we take a state-of-the-art syntax-based sentiment analysis system \citep{VilAloGom2016ArxivUUUSA}, which calculates the polarity of a text (i.e., whether it expresses a positive, negative or neutral stance) relying on its dependency parse tree; and we test it with a set of well-known syntactic parsers, including models with state-of-the-art accuracy and others that are less accurate, but have a smaller computational cost, evaluating how the choice of parser affects the accuracy of the polarity classification. Our results show that state-of-the-art parsing accuracy does not provide additional benefit for this sentiment analysis task, as all of the parsers tested produce similarly good polarity classification accuracy \changed{(no statistically significant differences, all p-values $\ge 0.49$)}. Therefore, our results suggest that it makes sense to use the fastest parsers for this task, even if they are not the most accurate.

The remainder of this article is organized as follows: we review the state of the art in syntactic parsing and syntax-based sentiment analysis in Section \ref{background}, we describe our experimental setup in Section \ref{methods}, we report the results in Section \ref{results}, and discuss their implications in Section \ref{discussion}. Finally, Section \ref{conclusions} draws our conclusion and discusses possible avenues for future work.

\section{Background}
\label{background}

We now provide an overview of research in parsing and sentiment analysis that is relevant to this study. 

\subsection{Parsing}
\label{parsing}

\begin{figure}[hbtp]
  \centering
  \includegraphics[width=9cm,clip]{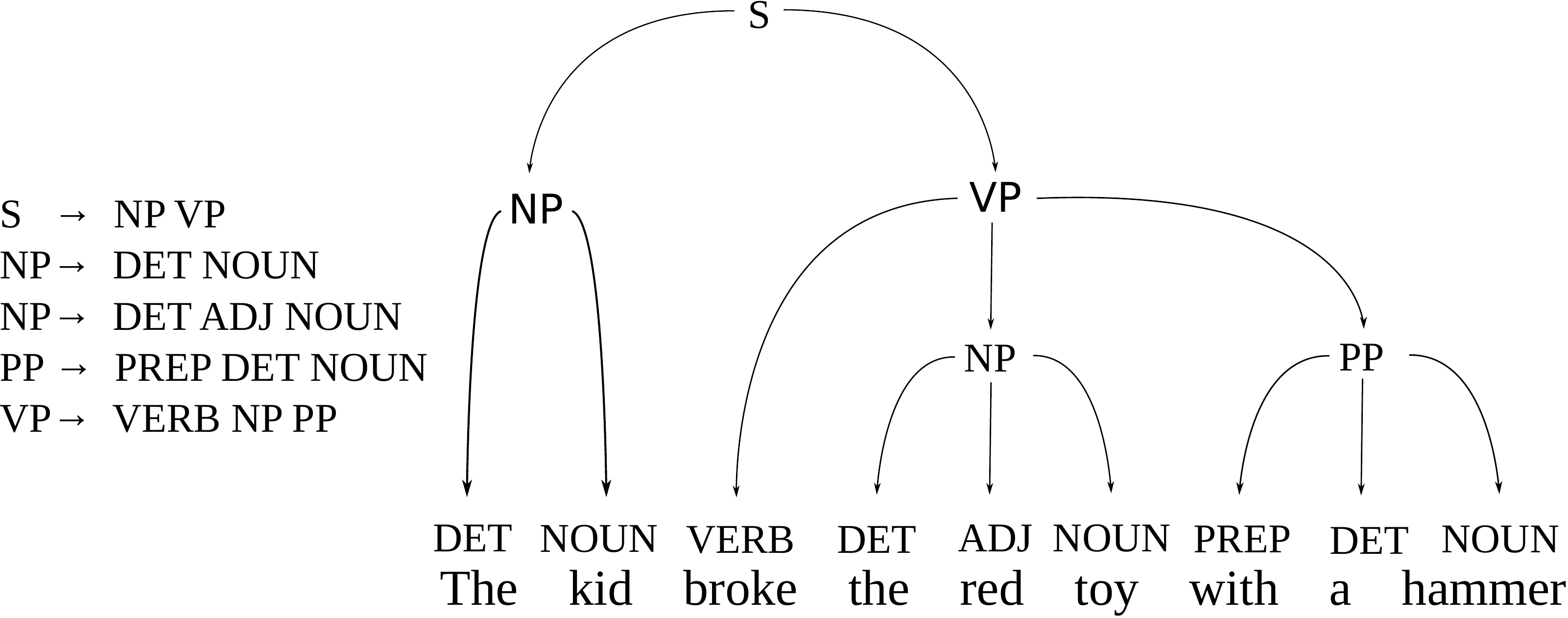}
  \caption{A valid constituency parse for the sentence \quotedtext{The kid broke the red toy with a hammer}. The sentence is divided into constituents according to the constituency grammar defined at the left part of the picture}
  \label{figure-constituency-example}
\end{figure}

\begin{figure}[hbtp]
  \centering
  \includegraphics[width=7cm,clip]{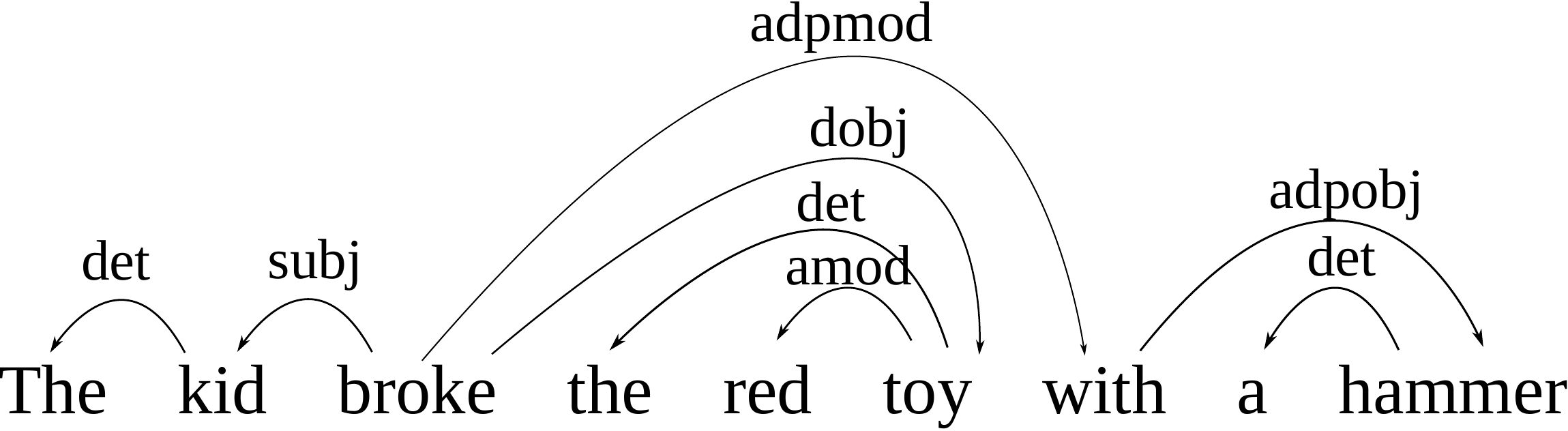}
  \caption{A valid dependency parse for the sentence: \quotedtext{The kid broke the red toy with a hammer}. The sentence is represented as a graph of binary relations between words that represent the existing syntactic relation between them (e.g. \quotedtext{kid} is the subject of the verb \quotedtext{broke})}
  \label{figure-dependency-example}
\end{figure}

Different linguistic theories define different ways in which the syntactic structure of a sentence can be described. In particular, the overwhelming majority of natural language parsers in the literature adhere to one of two dominant representations. In \emph{constituency grammar} (or \emph{phrase structure grammar}), sentences are analyzed by breaking them up into segments called constituents, which are in turn decomposed into smaller constituents, as in the example of Figure \ref{figure-constituency-example}. In \emph{dependency grammar}, the syntax of a sentence is represented by directed binary relations between its words, called dependencies, which are most useful when labeled with their syntactic roles, such as subject and object, as in Figure \ref{figure-dependency-example}. Each of these representation types provides different information about the sentence, and it is not possible to fully map constituency to dependency representations or vice versa \citep{Kahane2015polygraphs}.

In this paper we will focus on dependency parsing, as it is the predominant representation used by most of the downstream AI applications mentioned above -- with machine translation as an arguable exception, where constituent parsing is often used due to the adequacy of phrase structure grammar for modeling reordering of words between languages \citep{DeNeefe2009synchronous,Xiao2016translation} --, and it is also the alternative used by the syntax-based SA system we will use in our experiments.

Most dependency parsing systems in the literature can be grouped into two broad categories \citep{mcdonald07emnlp}: \emph{graph-based} and \emph{transition-based} (shift-reduce) parsers. 

Graph-based parsers use models that score dependency relations or groups of them, and perform a global search for a parse that will maximize the combined score of all dependencies. Under the assumption of projectivity (i.e., that there are no crossing dependencies), there are several dynamic programming algorithms that perform exact search in cubic time \citep{Eisner96,GomCarWeiACL2008}, but this restriction is not realistic in practice \citep{Gom2016restricted}. Unfortunately, exact inference has been shown to be intractable for models that support arbitrary non-projectivity, except under strong independence assumptions \citep{McDonald07} which enable parsing in quadratic time with maximum spanning tree algorithms \citep{McDonald05b}, but severely limit the expressivity of the feature models that can be used. This restriction can be avoided by using so-called \emph{mildly non-projective} parsing algorithms, which support the overwhelming majority of non-projective analyses that can be found in real 
linguistic structures \citep{GomCarWeiCL2011,CohGomSatEMNLP2011,Pitler2013}; but they have super-cubic complexities that make them too slow for practical use. Another option is to forgo exact inference, using approximate inference algorithms with rich feature models instead. This is the approach taken by TurboParser \citep{Martins2010turbo,Martins2013}, which currently is the most popular graph-based parser as it can provide state-of-the-art accuracy with a reasonable computational cost.

Transition-based parsers are based on a state machine that builds syntactic analyses step by step, typically from left to right. A statistical or machine learning model scores each of the possible transitions to take at each state, and a search strategy is used to find a high-scoring sequence of transitions. The earlier approaches to transition-based parsing, like the MaltParser system \citep{nivre07nle} used greedy deterministic search for this purpose, which is especially fast, but is prone to obtain suboptimal solutions due to bad decisions at early stages that result in error propagation. This problem is alleviated by instead performing beam search \citep{Zhang2011rich} or dynamic programming \citep{hs10,KuhGomSatACL2011} to explore transition sequences, but this increases the computational cost. Other alternatives that provide a good speed-accuracy tradeoff are selectional branching, which uses confidence estimates to decide when to employ a beam \citep{Choi2013}, or dynamic oracles, which reduce the 
error propagation in greedy search by exploring non-optimal transition sequences during training \citep{Goldberg2012dynamic}. In the last two years, several transition-based parsers have appeared that use neural networks as their scoring model \citep{CheMan2014,Dyer2015lstm,Andor2016face}, providing very good accuracy.


\subsection{Parsing Evaluation}

The standard metrics to evaluate the accuracy of a dependency parser are the unlabeled attachment score (UAS: the proportion of words that are attached to the correct head word by means of a dependency, regardless of its label), labeled attachment score (LAS: the proportion of words that are attached to the correct head by means of a dependency that has the correct label) and label accuracy (LA: the proportion of words that are assigned the correct dependency type). However, the performance of a parser in terms of such scores is not necessarily proportional to its usefulness for a given task, as not all dependencies in a syntactic analysis are equally useful in practice, or equally difficult to analyze \citep{NivRimMcDGomCOLING2010,Bender2011evaluation}. Therefore, LAS, UAS and LA are of limited use for researchers and practitioners that work with downstream applications in NLP. For this purpose, it is more useful to perform task-oriented evaluation, i.e., to experiment with the parsers in the actual tasks 
for which they are going to be used \citep{Volokh2012task}.

Such evaluations have been performed for some specific NLP tasks, namely information extraction \citep{Miyao2008task,Buyko2010evaluating}, textual entailment recognition \citep{Yuret2010evaluation,Volokh2012task} and machine translation \citep{Quirk2006impact,Goto2011comparison,Popel2011choice}. However, these comparisons are currently somewhat dated, as they were performed before the advent of the major advances in parsing accuracy of the current decade reviewed in Section \ref{parsing}, such as beam-search transition-based parsing, dynamic oracles, approximate variational inference (TurboParser) or neural network parsing. Even more importantly, these analyses provide very different results depending on each specific task and, to our knowledge, no evaluation of parsers has been performed for sentiment analysis, a task where a good speed-accuracy tradeoff is especially important due to its extensive applications to the Web and social networks.

\subsection{Syntax-based Sentiment Analysis}
\label{syntaxbased}

\changed{A number of state-of-the-art models for SA using different morphological \citep{khan2016esap,khan2016swims} and syntactic approaches have proven useful in recent years.} \cite{LiuGaoLiuZha2016a} pointed out the benefits of syntactical approaches with respect to statistical models on opinion target extraction, such as domain independence, and propose two approaches to select a set of rules, that even being suboptimal, achieve better results than a state-of-the-art conditional random field supervised method.
\cite{WuZhaHuaWu2009a} defined an approach to extract product features through phrase dependency parsing: they first combine the output of a shallow and a word-level dependency parser to then extract features and feed a support vector machine (SVM) with a novel tree kernel function. Their experimental results outperformed a number of bag-of-words baselines. \cite{JiaYuMen2009a} and \cite{asmi2012negation} defined a set of syntax-based rules for identifying and handling negation on natural language texts represented as dependency trees. They also pointed out the advantage of using this kind of methods with respect to traditional lexicon-based perspectives in tasks such as opinion mining or information retrieval.
\cite{PorCamWinHua2014a} posed a set of syntax-based patterns for a concept-level approach to determine how the sentiment flows from concept to concept, assuming that such concepts present in texts are represented as nodes of a dependency tree.

\cite{JosPen2009a} introduced the concept of generalized triplets, using them as features for a supervised classifier and showing its usefulness for subjectivity detection. Given a dependency triplet, the authors proposed to generalize the head or the dependent term (or even both at the same time) to its corresponding part-of-speech tag. Thus, the triplet \emph{(car, modified, good)} could be generalized as \emph{(NOUN, modifier, good)}, which can be useful to correctly classify similar triplets that did not appear in the training set (e.g.  \emph{(bicycle, modifier, good)} or \emph{(job, modifier, good)}).
In a similar line, \cite{VilAloGom2013d} enriched the concept of generalized dependency triplets and showed that they can be exploited as features to feed a supervised SA system for polarity classification, as long as enough labeled data is available.
The same authors \citep{VilAloGomNLE2015} proposed an unsupervised syntax-based approach for polarity classification on Spanish reviews represented as Ancora trees \citep{Ancora}. They showed that their system outperforms the equivalent lexical-based approach \citep{Lexicon-BasedMethods}. In this line, however, \cite{Lexicon-BasedMethods} pointed out that one of the challenges when using parsing techniques for sentiment analysis is the need of fast parsers that are able to process in real-time the huge amount of information shared by users in social media.

With the recent success of deep learning, \cite{SocPerWuChuManNgPot2013a} syntactically annotated a sentiment treebank to then train a recursive neural network that learns how to apply semantic composition for relevant phenomena in SA, such as negation or \quotedtext{but} adversative clauses, over dependency trees. \cite{KalGreBlu2014a} introduced a convolutional neural network for modeling sentences and used it for polarity classification among other tasks. Their approach does not explicitly rely on any parser, but the authors argue that one of the strengths of their model comes from the capability of the network to implicitly learn internal syntactic representations.

\section{Materials and Methods}
\label{methods}

We now describe the systems, corpora and methods used for our task-oriented evaluation.

\subsection{Parsing systems}
\label{section-parsing-systems}


\begin{itemize}
\item MaltParser: Introduced by \citet{nivre07nle}, this system can be used to train transition-based parsers with greedy deterministic search. Although its accuracy has fallen behind the state of the art, it is still widely used, probably owing to its maturity and solid documentation. Additionally, due to its greedy nature, MaltParser is very fast. Following common practice, we use it together with the feature optimization tool MaltOptimizer\footnote{MaltParser often requires feature optimization to obtain acceptable results for the target language.} \citep{MaltOptimizer} to optimize the parameters and train a suitable model. The trained MaltParser model uses a standard arc-eager (transition-based) parsing algorithm, where at each step the movement to apply is selected among the set of possible transitions, previously scored by a linear model, which is faster than using models based on SVMs.
\item TurboParser \citep{Martins2013}: A graph-based parser that uses approximate variational inference with non-local features. It has become the most widely used graph-based parser, as it provides better speed and accuracy than previous alternatives. We use its default configuration, training  a second-order non-projective parser with features for arcs, consecutive siblings and grandparents, using the AD3 algorithm as a decoder.

\item YaraParser \citep{Rasooli2015yara}: A recent transition-based parser, which uses beam search \citep{Zhang2011rich} and dynamic oracles \citep{Goldberg2012dynamic} to provide state-of-the-art accuracy. Its default configuration is used.
\item Stanford RNN Parser \citep{CheMan2014}: The most popular among the recent wave of transition-based parsers that employ neural networks, it can achieve robust accuracy in spite of using greedy deterministic search. We use pretrained GloVe \citep{pennington2014glove} embeddings as input to the parser: in particular, 50-dimensional word embeddings\footnote{http://nlp.stanford.edu/data/glove.6B.zip} trained on Wikipedia and the English Gigaword \citep{napoles2012annotated}.
\end{itemize}

\subsection{Parsing corpus}
\label{section-parsing-treebanks}

To train and evaluate the parsing accuracy of such parsers, we are using the English Universal Treebank v2.0 created by \cite{McdNivQuirGolDasGanHalPetZhaTacBedCasLee2013}. It is a mapping from the (constituency) Penn treebank \citep{marcus1993building} to a universal dependency grammar annotation. The choice of the treebank is due to the already existing predefined compositional operations in the SA system used for evaluation (see \S \ref{section-sentiment-analysis-system}), that are intended for this type of universal guidelines. The corpus contains 39\,833, 1\,701 and 2\,416 dependency trees for the training, development and test sets, respectively, and it represents one of the largest available treebanks for English.

\subsection{Sentiment analysis system}
\label{section-sentiment-analysis-system}

For the task-oriented evaluation, we will rely on UUUSA, the universal, unsupervised, uncovered approach for sentiment analysis described by \citep{VilAloGom2016ArxivUUUSA}, which is based on syntax and the concept of \emph{compositional operations}. Briefly, given a text represented as a dependency tree, a \emph{compositional operation} defines how a node of the tree modifies the semantic orientation (a real value representing a polarity and its strength) of a different branch or node, based on features such as its word form, part-of-speech tag or dependency type, without any limitation in terms of its location inside such tree. The associated system queues operations and propagates them through the tree, until the moment they must be dequeued and applied to their target. The model has outperformed other state-of-the-art lexicon-based methods on a number of corpora and languages, showing the advantages of using syntactic information for sentiment analysis. Due to the way the system works, in such a way that 
the application of the operation relies on previously assigning dependency types and heads correctly, it also constitutes a proper environment to test how parsing accuracy affects polarity classification.

The system already includes a predefined set of universal syntactic operations, that we are using in this study to determine the importance of parsing accuracy. For the sake of brevity, we are not detailing how the system computes the semantic orientation of the trees, but we specify which universal dependencies UUUSA is relying on to identify relevant linguistic phenomena that should trigger a compositional operation. To apply an operation, usually a dependency type must match at the node a branch is rooted at. The existing set of predefined operations that we are considering involve phenomena such as:
\begin{itemize}
 \item \emph{Intensification}: A branch amplifies or decreases the semantic orientation of its head node or other branch (that must be labeled with the \emph{acomp} (adjectival complement) dependency type). The intensifier branch must be labeled as one of these three dependency types: \emph{advmod} (adverb modifier), \emph{amod} (adjective modifier), \emph{nmod} (noun modifier). Dependencies are relevant in this case because they help avoid false positive cases when applying intensification (e.g. in \quotedtext{It is huge}, \quotedtext{huge} should be (probably) a positive adjective, meanwhile in \quotedtext{I have huge problems} it acts as an intensifier as it is an adjective modifier of the negative word \quotedtext{problems}, and in \quotedtext{I have huge exciting news} it acts again as an intensifier, but of a positive term).
 \item \quotedtext{But} \emph{clauses}: To trigger this compositional operation, which decreases the relevance of the semantic orientation of the main sentence, the dependent branch rooted at \emph{\textquoteleft but\textquoteright} must be labeled as \emph{cc}.
 \item \emph{Negation}: The negating terms, that might shift the sentiment of other branches, are labeled in a dependency tree with the dependency type \emph{neg}.
 \item \quotedtext{If}:  We also include experiments using the proposed rule in \cite{VilAloGom2016ArxivUUUSA} for the \quotedtext{if} clause, which is labeled with the \emph{mark} dependency type, assuming that the part of the sentence under the scope of influence of the conditional clause should be ignored.
\end{itemize}

Therefore, the accuracy obtained by UUUSA on the sentiment corpora is related to the parsing accuracy: a LAS of zero makes it impossible to trigger any compositional operation, since no dependency type would match; obtaining as output a global polarity which is the result of simply summing the semantic orientation of individual words.  

\begin{figure}[hbtp]
  \centering
  \includegraphics[width=8.5cm,clip]{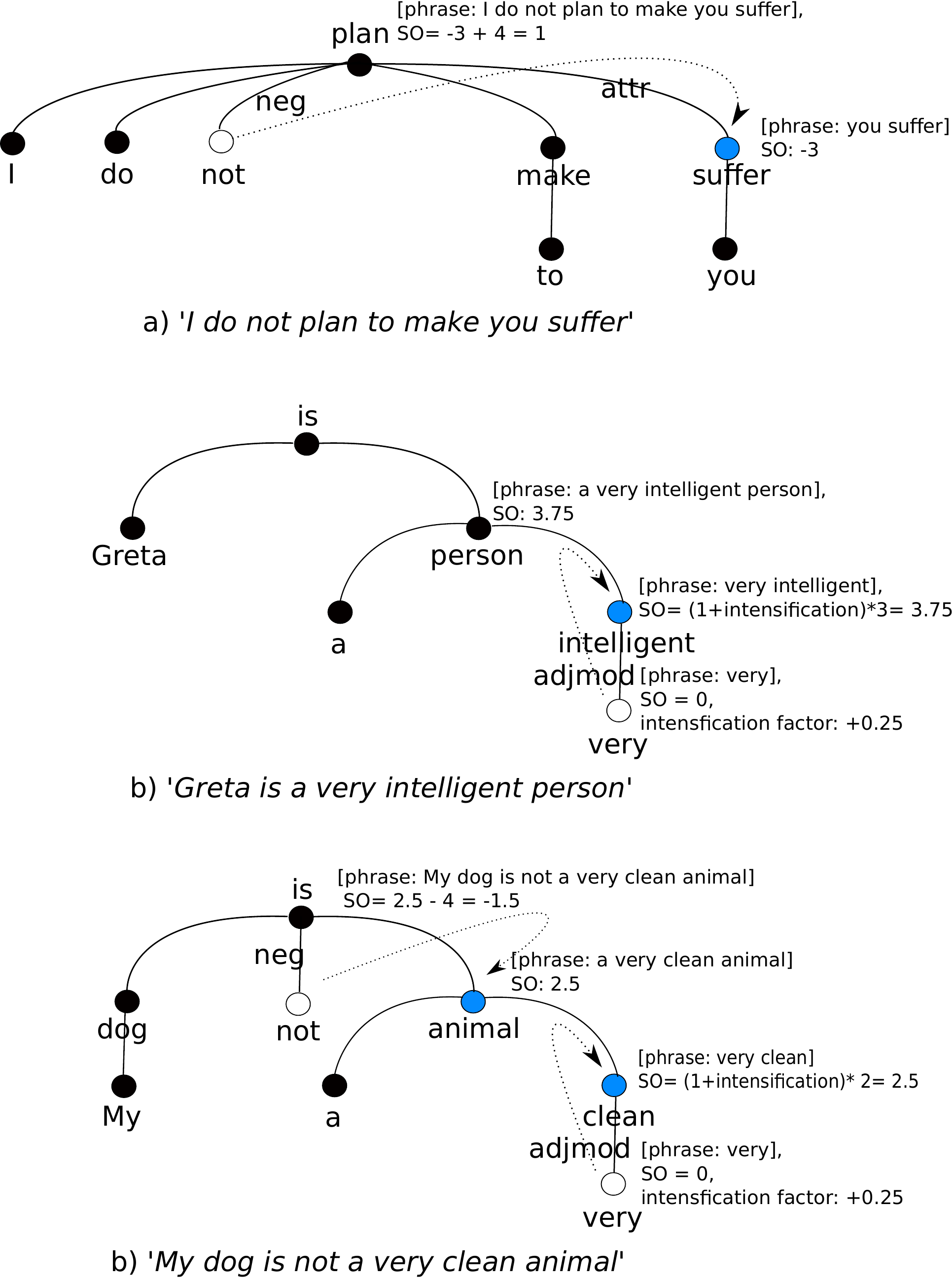}
  \caption{Analysis of three sentences using the UUUSA approach. The analysis corresponds to a post-order recursive traversal.
  Semantic orientation, intensification and negation values are orientative. At each level, we show for the relevant nodes, those playing any role in the computation of the semantic orientation, the corresponding phrase rooted at that node, and its corresponding semantic orientation (SO), once compositional operations have been applied at that level. Sentence a) shows an example where the semantic scope of the negation is non-local, but thanks to dependency parsing and syntactic rules, the system can accurately identify such scope and shift the semantic orientation coming from that branch. Note that, if either the dependency type \emph{neg} or \emph{attr} were assigned incorrectly, the calculation of the SO would be wrong. Sentence b) illustrates how the term \emph{\textquoteleft very\textquoteright} increases the semantic orientation of its head. It is important to remark that if the dependency type \emph{adjmod} were assigned incorrectly, the analysis would be again unaccurate.
  Sentence c) illustrates a more complex compositional example, where first, the intensifier \emph{\textquoteleft very\textquoteright} amplifies the semantic orientation of the word \emph{\textquoteleft clean\textquoteright}, and the negating word \emph{\textquoteleft not\textquoteright} shifts the sentiment rooted at the phrase \emph{\textquoteleft a very clean animal\textquoteright}}
  \label{figure-running-example}
\end{figure}

Figure \ref{figure-running-example}.a) and Figure \ref{figure-running-example}.b) illustrate two simple examples where part-of-speech tags, dependencies and types play a relevant role to accurately capture the semantic orientation of the sentence. Additionally, Figure \ref{figure-running-example}.c) illustrates with an additional example how semantic composition is managed when a negation and an intensification appear in the same sentence and affect the same subjective word.

 \changed{We chose this system among others for three main reasons:}
 \begin{enumerate}
  \item \changed{It supports separate compositional operations to address very specific linguistic phenomena, which can be enabled or disabled individually. This gives us great flexibility to carry out experiments including and excluding a number of linguistic constructions, allowing us to determine how relevant parsing accuracy is to tackle each of them.}
  \item \changed{It is a modular system where the parser is an independent component that can be swapped with another parser, allowing us to use it for task-oriented evaluation of various parsers. This contrasts with Socher et al (2013), a system that also uses syntax, but where the parsing process is tightly woven with the sentiment analysis process (a neural network architecture is trained to perform both tasks at the same time) so that it is not possible to use it with the output of external parsers.}
  \item \changed{Symbolic or knowledge-based systems like this perform robustly across different datasets and domains, which we cannot guarantee for the case of many machine learning models, that do not generalize so well (Aue and Gamon 2005; Taboada et al 2011; Vilares et al 2017).}

 \end{enumerate}

\subsection{Sentiment analysis corpora}

Three standard corpora for document- and sentence-level sentiment analysis are used for the extrinsic evaluation:

\begin{itemize}
 \item \cite{taboada2004analyzing} corpus: A general-domain dataset composed of 400 long reviews (50\% positive, 50\% negative) about different topics (e.g. washing-machines, books or computers). 
 \item \cite{PangBoandLee2004} corpus: A collection of 2\,000 long movie reviews (50\% positive, 50\% negative). 
 \item \cite{PanLee2005} corpus: A collection of short (i.e. single-sentence) movie reviews. We relied on the test split used by \cite{SocPerWuChuManNgPot2013a}, removing the neutral ones, as they did, for the binary classification task (1\,821 subjective sentences: $\sim$\,49\% positive, $\sim$\,51\% negative).
\end{itemize}

\subsection{Experimental methodology}

The aim of our experiments is to show how parsing accuracy influences polarity classification, following a task-oriented evaluation. To do so, we first compare the performance of different parsers  on a standard treebank test set and metrics. We then extrinsically evaluate the performance of such parsers by parsing sentiment corpora, and using the obtained parse trees to determine the polarity of the texts in the corpora by means of a state-of-the-art syntax-based model. The performance of this model relies on previous correct assignment of dependency types and heads, to be able to handle relevant linguistic phenomena for the purpose at hand (e.g. intensification, \quotedtext{but} clauses or negation). This makes it possible to relate parsing and syntax-based sentiment performance.

\subsection{\changed{Hardware and software used in the experiments}}

\changed{Experiments were carried in a Dell XPS 8500 Intel Core i7 @ 3.4GHz and 16GB of RAM. Operating system was Ubuntu 14.04 64 bits.}

\section{Results}
\label{results}

Table \ref{tab:ut-las-uas} shows the performance obtained by the different parsers according to the standard metrics: LAS, UAS and LA. Table \ref{tab:ut-time} illustrates how much time each parser consumes to analyze the \cite{PanLee2005} corpus, and the total time once the SA system is run on it. 

Tables \ref{tab:pl04-acc}, \ref{tab:pl05-acc} and \ref{tab:sfu-acc} show the accuracy obtained by UUUSA on different sentiment corpora, when the output of each of the parsers is used as input to the syntax-based sentiment analysis system.\footnote{The results obtained in these corpora are slightly different from the ones reported by \cite{VilAloGom2016ArxivUUUSA}, due to the different tokenization techniques used in this work.}
We take accuracy as the reference metric for the SA systems, because it is the most suitable metric in this case, since the three corpora are balanced. In particular, we compare the performance when no syntactic rules are used (which would be equivalent to a lexicon-based system that only sums the semantic orientation of individual words), with respect to the one obtained when different rules are added. The aim is to determine if different parsers manage relevant linguistic phenomena in a different way.
 
\begin{figure*}[hbtp]
  \centering
  \includegraphics[width=14.0cm,clip]{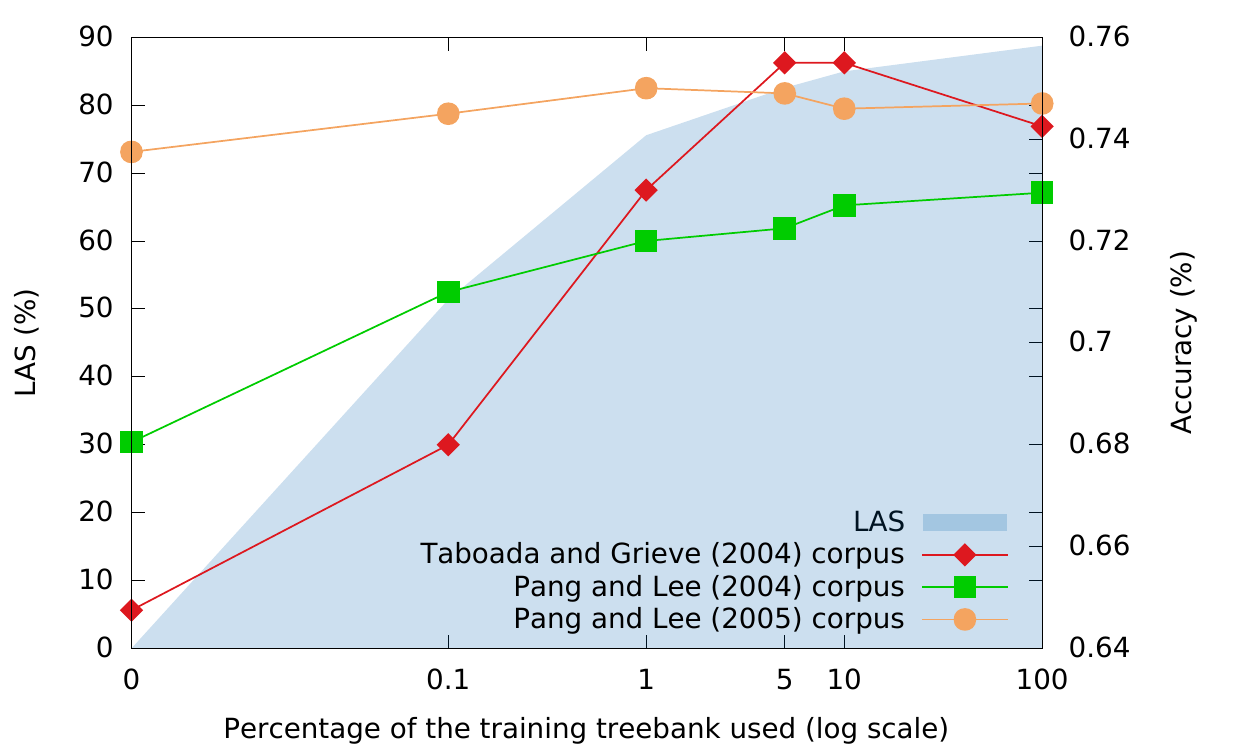}
  \caption{Relationship between LAS (area graphic, left y-axis) and accuracy in different sentiment corpora (line graphics, right y-axis), using the Stanford RNN parser \citep{CheMan2014} trained with different portions (\%) of the training treebank (x-axis). The plot shows that using a larger training treebank improves LAS, but does not necessarily increase the UUUSA sentiment accuracy, especially when more than a 5 or 10\% of such treebank is used to train such parser.}
  \label{figure-relation-las-acc}
\end{figure*}

Finally, Figure \ref{figure-relation-las-acc} relates the LAS performance on the test set of the universal treebank with respect to the accuracy obtained by UUUSA, when we artificially reduce the training set size to simulate a low-accuracy parsing setting, as could happen in low-resource languages.

%

\begin{table}[hbtp]
        \centering
	\caption{Performance (LAS, UAS and LA)  of the parsers on the English Universal treebank test set. We also detail the performance in terms of Precision (P) and Recall (R) for the dependency types that are playing a role in the predefined compositional operations of UUUSA. The subscripts indicate the rank of the parser with respect to the others, given a particular metric}
	\label{tab:ut-las-uas}       
	\begin{tabular}{|ccccc|}
		\hline
		\bf Metric&\bf MaltParser &\bf Stanford RNN parser &\bf TurboParser &\bf YaraParser \\
		\hline
		\hline
		\bf LAS & 88.35$_4$ & 88.77$_3$ & 91.36$_2$  &\bf 91.84$_1$  \\
		\bf UAS & 90.27$_4$ & 90.47$_3$ & 93.29$_2$  &\bf 93.34$_1$   \\
		\bf LA & 93.01$_4$  & 93.59$_3$ & 95.02$_2$  &\bf 95.72$_1$  \\
		\hline
		\hline
		\bf P(\emph{acomp}) & 88.66$_3$ & 88.31$_4$ & 88.75$_2$ &\bf 91.03$_1$\\
		\bf R(\emph{acomp}) & 90.29$_3$ & 89.24$_4$ & 91.08$_2$ &\bf 93.18$_1$\\
		\hline
		\hline
		\bf P(\emph{advmod}) & 83.18$_3$ & 82.38$_4$ & 84.96$_2$ &\bf 85.85$_1$\\
		\bf R(\emph{advmod}) & 84.04$_3$ & 81.97$_4$ &\bf 85.46$_1$ & 85.22$_2$\\	
		\hline
		\hline
		\bf P(\emph{amod})  & 95.45$_3$ & 95.01$_4$ &\bf 96.25$_1$ & 96.23$_2$\\		
		\bf R(\emph{amod})  & 95.45$_3$ & 95.40$_4$ & 96.30$_2$ &\bf 96.60$_1$\\		
		\hline
		\hline
		\bf P(\emph{attr}) & 87.17$_3$ & 86.00$_4$ & 89.00$_2$ &\bf 94.88$_1$\\
		\bf R(\emph{attr}) & 88.63$_3$ & 86.29$_4$ & 91.97$_2$ &\bf 92.98$_1$\\
		\hline
		\hline
		\bf P(\emph{cc}) & 77.06$_4$ & 77.68$_3$ &\bf 83.56$_1$ & 83.46$_2$\\
		\bf R(\emph{cc}) & 76.95$_4$ & 77.02$_3$ & 83.82$_2$ &\bf 83.97$_1$\\
		\hline
		\hline
		\bf P(\emph{mark})  & 83.44$_4$ & 84.21$_3$ & 85.15$_2$ &\bf 88.47$_1$ \\
		\bf R(\emph{mark})  & 83.44$_4$ & 88.31$_3$ & 90.26$_2$ &\bf 92.21$_1$\\
		\hline
		\hline
		\bf P(\emph{neg}) & 92.99$_2$  & 92.62$_4$  &\bf 94.77$_1$  & 92.68$_3$ \\
		\bf R(\emph{neg}) & 94.72$_2$  & 93.48$_4$  &\bf 95.65$_1$  & 94.41$_3$  \\
		\hline
		\hline
		\bf P(\emph{nmod})  & 80.66$_4$ & 82.17$_3$ & 84.45$_2$ &\bf 85.38$_1$ \\
		\bf R(\emph{nmod}) & 79.67$_2$ & 78.66$_4$ & 79.47$_3$ &\bf 80.69$_1$\\		
		\hline
	\end{tabular}
\end{table}


\begin{table}[hbtp]
\centering
\caption{Average, maximum and minimum execution time (seconds) out of 5 runs on the \citet{PanLee2005} test set. We also include the total execution time, after the SA system has been run on the \cite{PanLee2005} corpus}
\label{tab:ut-time}       
\begin{tabular}{|l|cccc|}
\hline
\bf Parser&\bf Average &\bf Minimum &\bf Maximum&\bf Average + UUUSA \\
          &            &            &           &\bf time ($\sim$\,1.2 sec)\\
	\hline
	\hline
	MaltParser &\bf 4.02 &\bf 3.83 &\bf 4.23&\bf 5.22\\
	Stanford RNN Parser  & 5.99 & 5.82 & 6.23 & 7.19\\
	Turbo Parser & 97.34 & 94.79 & 99.23& 98.54 \\
	Yara Parser  & 39.85 & 38.94 & 41.23 & 41.05\\
	\hline
	\end{tabular}
\end{table}

\newpage

\begin{table}[hbtp]
	\caption{Accuracy on the \cite{PangBoandLee2004} corpus considering different subsets of rules}
	\label{tab:pl04-acc}       
	\begin{tabular}{|c|cccccc|}
	        \hline
		\bf Parser &\bf All &\bf None &\bf Intensification &\bf \quotedtext{but} &\bf \quotedtext{if} &\bf Negation \\
		\hline
		\hline
		MaltParser  & 72.75 & 68.05 & 70.35 & 67.75 & 68.20 & 69.80 \\
		Stanford RNN Parser &\bf 72.95 & 68.05 &\bf 70.60 & 67.85 & 68.35 &\bf 70.30 \\
		Turbo Parser & 72.20 & 68.05 & 70.35 &\bf 67.95 & 68.25 & 69.60 \\
		Yara Parser & 72.00 & 68.05 & 70.30 & 67.85 &\bf 68.55 & 70.15 \\
		\hline
	\end{tabular}
\end{table}

\begin{table}[hbtp]
	\caption{Accuracy on the \cite{PanLee2005} corpus considering different subsets of rules}
	\label{tab:pl05-acc}       
	\begin{tabular}{|c|cccccc|}
		\hline
		\bf Parser &\bf All &\bf None &\bf Intensification &\bf \quotedtext{but} &\bf \quotedtext{if} &\bf Negation \\
		\hline
		\hline
		MaltParser & \bf 74.79 & 73.75 &\bf 74.41 & 73.86 & 73.75 & 74.14 \\
		Stanford RNN Parser  & 74.68 & 73.75 & 74.35 & 73.86 &\bf 73.97 &\bf 74.19 \\
		Turbo Parser & 74.57 & 73.75 &\bf 74.41 &\bf 73.92 & 73.86 &\bf 74.19 \\
		Yara Parser & 74.68 & 73.75 &\bf 74.41 & 73.86 &\bf 73.97 & 73.92 \\
		\hline
	\end{tabular}
\end{table}

\begin{table}[hbtp]
	\caption{Accuracy on the \cite{taboada2004analyzing} corpus considering different subsets of rules}
	\label{tab:sfu-acc}       
	\begin{tabular}{|c|cccccc|}
		\hline
		\bf Parser &\bf All &\bf None &\bf Intensification &\bf \quotedtext{but} &\bf \quotedtext{if} &\bf Negation \\
		\hline\hline
		MaltParser  & 74.00 & 64.75 & 66.25 &\bf 64.75 & 64.00 & 72.25 \\
		Stanford RNN Parser & 74.25 & 64.75 & 66.25 &\bf 64.75 &\bf 64.25 &\bf 72.75 \\
		Turbo Parser &\bf 75.00 & 64.75 &\bf 66.50 & 64.50 & 63.75 &\bf 72.75 \\
		Yara Parser & 73.50 & 64.75 &\bf 66.50 & 64.50 & 63.75 & 72.00 \\
		\hline
	\end{tabular}
\end{table}

\section{Discussion}
\label{discussion}

The results illustrated in Tables \ref{tab:ut-las-uas} and \ref{tab:ut-time} indicate the relationship between the parsing time and accuracy. The slower parsers \citep{Martins2013,Rasooli2015yara} tend to obtain a better performance, meanwhile the faster ones \citep{nivre07nle,CheMan2014} attain worse LAS, UAS and LA. This fact is expected, as there is a well-known tradeoff between speed and accuracy in the spectrum of parsing algorithms, with one extreme at greedy search approaches that scan and parse the sentence in a single pass but are prone to error propagation, and the other at exact search algorithms that guarantee finding the highest-scoring parse under a rich statistical model, but are prohibitively slow \citep{Choi2013,VolokhPhD,Gom2016restricted}. The tendency remains when looking at the performance on individual dependency types (where also the head is assigned correctly). 

However, a better LAS or UAS does not necessarily translate into a higher sentiment accuracy, which is shown in Tables \ref{tab:pl04-acc}, \ref{tab:pl05-acc} and \ref{tab:sfu-acc}. In most cases, the performance obtained by the different parsers under the same sets of rules is practically equivalent. \changed{To confirm this statistically, we applied chi-squared significance tests to compare the outputs obtained using the different parsers for each given dataset and set of sentiment rules. No significant differences in sentiment accuracy were found in any of these experiments, which reinforces our conclusion. The minimum p-value obtained was 0.49.} It is important to remark that this is very different from stating that parsing is not relevant for SA. In the case of UUUSA, \cite{VilAloGom2016ArxivUUUSA} already showed that their syntax-based SA approach is able to beat purely lexicon-based methods on a number of languages. In this line, Tables \ref{tab:pl04-acc}, \ref{tab:pl05-acc} and \ref{tab:sfu-acc} also show that the sets of syntactic rules outperform the baseline that does not use any syntactic-based rules (\textquoteleft None\textquoteright\ column) in almost all cases, proving again that syntax-based rules are useful to handle relevant linguistic phenomena in the field of SA. 

\changed{
The specific reasons that explain why the choice of syntactic parsing algorithm does not significantly affect accuracy lie out of the scope of our empirical work, as they require an exhaustive linguistic analysis. In view of the data, possible factors that may contribute are the following:
\begin{itemize}
\item Low difficulty of some of the most decisive dependencies involved: as can be seen in Table 1, even the least accurate parsers analyzed are obtaining well over 92\% precision and recall in adjectival modifiers (amod) and negations (neg), which are crucial for handling intensification and negation. This is likely because these tend to be short-distance dependencies, which are easier to parse (McDonald and Nivre 2007), and are common so they do not suffer from training sparsity problems. Thus, a highly accurate parser is not needed to detect these particular dependencies correctly.
\item Redundancy in sentences: a sentence may include several expressions of sentiment, so that even if the parse tree contains inaccuracies in a part of the sentence, we may still be able to extract the correct sentiment from the rest. This can be especially frequent in long sentences, which are the most difficult to parse (McDonald and Nivre 2007).
\item Irrelevance of fine-grained distinctions: in some cases, the parser provides more information than is strictly needed to evaluate the sentiment of a sentence. For example, the UUUSA rule for intensifiers works in the same way for adverbial modifiers (advmod), adjectival modifiers (amod) or nominal modifiers (nmod). Thus, if a parser mistakes e.g. an advmod for an amod, this counts as a parsing error, but has no influence in the sentiment output.
\end{itemize}
However, verifying and quantifying the influence of each of these factors remains as an open question, which we would like to explore in the near future.}

An interesting conclusion that could be extracted from these results is that parsing should prioritize speed over accuracy for syntax-based polarity classification. We draw Figure \ref{figure-relation-las-acc} to reinforce this hypothesis. The figure illustrates how LAS and sentiment accuracy vary when training the Stanford RNN parser \citep{CheMan2014} with different training data size. To do so, we trained a number of parsers using the first $x\%$ of the training treebank. As expected, it was observed that adding more training data increased the LAS obtained by the parser. However, this same tendency did not remain with respect to sentiment accuracy, which remains stable once LAS reaches an acceptable level. Based on empirical evaluation, sentiment accuracy stops increasing when using the first 5\% (82.57\% LAS) or 10\% (84.99\% LAS) of the English Universal training treebank, with which it is possible to already obtain a performance close to the state of the art (88.77\% when using the whole training 
treebank). On the other hand, there is a clear increasing tendency when $x < 5$, because in those cases the LAS is still not good enough (using the first 0.1\% and 1\% of the training treebank we only are able to achieve a LAS of 51.39\% and 75.58\%, respectively).
%

%
%

\section{Conclusions}\label{conclusions}

In this article, we have carried out a task-oriented empirical evaluation to determine the relevance of parsing accuracy on the primary challenge of sentiment analysis: polarity classification. We chose English as the target language and trained a number of standard and freely available parsers on the Universal Dependency Treebank  v2.0 \citep{McdNivQuirGolDasGanHalPetZhaTacBedCasLee2013}. The output of such parsers on different standard sentiment corpora is then used as input for a state-of-the-art and syntax-based system that aims to classify the polarity of those texts. Experimental results let us draw two interesting and promising conclusions: (1) a better labeled/unlabeled attachment score on parsing does not necessarily imply a significantly better accuracy on polarity classification when using syntax-based algorithms and (2) parsing for sentiment analysis should focus on speed instead of accuracy, as a LAS of around 80\% (which we obtained in the experiments by using only the first 10\% of the 
training treebank) is already good enough to fully take advantage of dependency trees and exploit syntax-based rules. Using larger training portions produces increases in the labeled attachment score up to the maximum value of close to 92\% that we obtained with the most accurate parser, but the performance for sentiment accuracy remains stable. Hence, there is no reason to use a slower parser to maximize LAS as long as one is above said \textquotedblleft good enough\textquotedblright\ threshold for sentiment analysis, which is clearly surpassed by all the parsers tested.

Based on the results, we believe there is room for improvements. We plan to design algorithms for faster parsing \citep{VolokhPhD}, prioritizing speed over accuracy. We also would like to explore the influence of parsing accuracy on other high-level tasks analysis, such as aspect extraction \citep{WuZhaHuaWu2009a} or question answering \citep{SquadArxiv}, where dependencies have played an important role.

%
%


\bibliographystyle{spbasic}      
\bibliography{carlos-own,twoplanaracl,evaluation-parsers-OM,main}   


\end{document}